\documentclass[sigconf]{acmart}
\usepackage{multirow}
\usepackage{multicol}
\usepackage{caption}

\AtBeginDocument{%
  \providecommand\BibTeX{{%
    \normalfont B\kern-0.5em{\scshape i\kern-0.25em b}\kern-0.8em\TeX}}}

\usepackage{natbib}
\usepackage[resetlabels]{multibib}
\newcites{sec}{References}

\newcites{latex}{\LaTeX-Literature}

\copyrightyear{2021}
\acmYear{2021}
\setcopyright{acmcopyright}\acmConference[MM '21]{Proceedings of the 29th ACM International Conference on Multimedia}{October 20--24, 2021}{Virtual Event, China}
\acmBooktitle{Proceedings of the 29th ACM International Conference on Multimedia (MM '21), October 20--24, 2021, Virtual Event, China}
\acmPrice{15.00}
\acmDOI{10.1145/3474085.3475504}
\acmISBN{978-1-4503-8651-7/21/10}

\acmSubmissionID{1822}

\begin{document}
\fancyhead{}

\title{Improving Robustness and Accuracy via Relative Information Encoding in 3D Human Pose Estimation}


\renewcommand\footnoterule{\kern-3pt \hrule width 2in \kern 2.6pt}


\author{Wenkang Shan$^{1}$,\quad Haopeng Lu$^{2}$,\quad Shanshe Wang$^{1,4}$*,\quad Xinfeng Zhang$^{3}$,\quad Wen Gao$^{1}$}

\makeatletter
\def\authornotetext#1{
\if@ACM@anonymous\else
    \g@addto@macro\@authornotes{
    \stepcounter{footnote}\footnotetext{#1}}
\fi}
\makeatother
\authornotetext{Corresponding author.}

\affiliation{
 \institution{\textsuperscript{\rm 1}Institute of Digital Media, Peking University, Beijing, China}
 \institution{\textsuperscript{\rm 2}Shanghai Jiao Tong University, Shanghai, China}
 \institution{\textsuperscript{\rm 3}University of Chinese Academy of Sciences, Beijing, China}
 \institution{\textsuperscript{\rm 4}Information Technology R\&D Innovation Center, Peking University, Shaoxing, China}
  }
\email{{wkshan,sswang,wgao}@pku.edu.cn, luhp2018@sjtu.edu.cn, xfzhang@ucas.ac.cn}

\def\authors{Wenkang Shan, Haopeng Lu, Shanshe Wang, Xinfeng Zhang, Wen Gao}










\begin{abstract}
Most of the existing 3D human pose estimation approaches mainly focus on predicting 3D positional relationships between the root joint and other human joints (local motion) instead of the overall trajectory of the human body (global motion). Despite the great progress achieved by these approaches, they are not robust to global motion, and lack the ability to accurately predict local motion with a small movement range. To alleviate these two problems, we propose a relative information encoding method that yields positional and temporal enhanced representations. Firstly, we encode positional information by utilizing relative coordinates of 2D poses to enhance the consistency between the input and output distribution. The same posture with different absolute 2D positions can be mapped to a common representation. It is beneficial to resist the interference of global motion on the prediction results. Second, we encode temporal information by establishing the connection between the current pose and other poses of the same person within a period of time. More attention will be paid to the movement changes before and after the current pose, resulting in better prediction performance on local motion with a small movement range. The ablation studies validate the effectiveness of the proposed relative information encoding method. Besides, we introduce a multi-stage optimization method to the whole framework to further exploit the positional and temporal enhanced representations. Our method outperforms state-of-the-art methods on two public datasets. Code is available at https://github.com/paTRICK-swk/Pose3D-RIE.

\end{abstract}




\maketitle

\section{Introduction}
3D human pose estimation aims to localize the positions of human joints in 3D space from a given RGB image or video. It is an active research topic in computer vision since it can be used in a broad range of domains, such as object detection \cite{wan2020faster}, video surveillance \cite{liu2018pose}, and augmented reality \cite{lin2010augmented}. 3D human pose estimation is a challenging ill-posed task due to the ambiguity that multiple 3D poses can have the same projection in 2D space. Traditional methods \cite{belagiannis20143d,andriluka2009pictorial,agarwal20043d} extend Pictorial Structure (PS) model to 3D pose estimation or utilize hand-crafted features to recover human poses.  With the help of 3D large-scale motion capture datasets \cite{ionescu2013human3,sigal2010humaneva}, learning-based methods \cite{fabbri2020compressed,sun2017compositional,pavlakos2017coarse}, which utilize neural networks to regress the 3D coordinates of human joints, have achieved promising results.

Recent approaches \cite{martinez2017simple,zhao2019semantic,jllo20193d} follow the two-step principle for 3D pose reasoning. The first step is to localize 2D human keypoints, while the second step is to predict the corresponding 3D joint locations from the results of the previous step. These approaches focus on the process of lifting 2D keypoints to a 3D skeleton. In this way, the prior information of 2D coordinates can be utilized to facilitate 3D human pose estimation. Our work falls under this category. 

\begin{figure}
  \centering
  \includegraphics[width=0.7\linewidth]{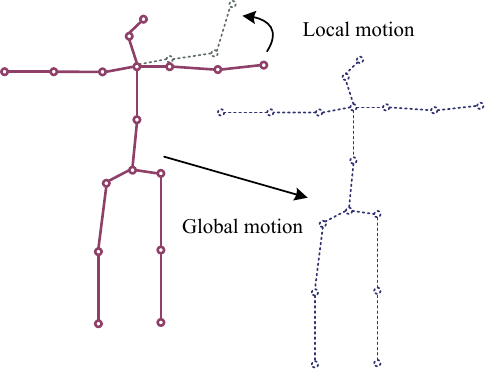}
  \vspace{-0.3cm}
  \caption{Illustration of the global and local motion. Global motion is the overall offset of all human joints. Local motion is the movement of each joint with respect to the root joint.}
  \vspace{-0.3cm}
  \label{img10}
\end{figure}

As illustrated in Figure~\ref{img10}, the motion of the human body can be divided into global and local motion. Most of the previous works \cite{martinez2017simple,liu2020attention,cheng2019occlusion,sun2017compositional,cai2019exploiting,liu2020attention} pay attention to the prediction of local motion instead of global motion. They represent 3D human joints in the form of relative coordinates with respect to the root joint (i.e. pelvis). Despite the great success achieved by these approaches, there remain two drawbacks. First, existing methods lack robustness to global motion. They only use the absolute positions of 2D poses as inputs, resulting in a gap between the distribution of the input and output. In real cases, the camera position often shifts, which can be regarded as the global motion of 2D keypoints. This brings a serious problem that the same posture of a person with different absolute 2D positions will correspond to different outputs. In other words, prediction results are affected by the movement of the camera. Second, previous works \cite{jllo20193d,liu2020attention} produce inaccurate prediction results on local motion. They use 2D coordinates within a period of time to predict the 3D pose of a person in a certain frame. These works treat every pose equally, ignoring the relationships between the current pose and all the others. This indicates that the network is insensitive to small changes in local motion, resulting in poor prediction performance on poses with a small movement range.

In this paper, we propose a relative information encoding method to alleviate the aforementioned problems. We begin with a baseline called Feature Fusion Network, utilizing the human body grouping strategy \cite{park20183d}. The relative information encoding produces positional and temporal enhanced representations as additional inputs to the baseline. First, we encode the positional information by using relative 2D coordinates with respect to the root joint at the input to ensure consistency with the output. In this way, the posture-related information can be distilled without being disturbed by the absolute position of the human body. When the positions of 2D keypoints in the image plane are globally shifted, the results after positional information encoding is still the same. Therefore, the 3D pose prediction becomes more robust to global motion. Second, to achieve a better estimation performance on local motion with a small movement range, we encode the temporal information by explicitly propagating the influence of the current pose to other poses. The temporal information encoding can be modeled as any vector operator, such as inner-product and subtraction. The position changes in the contextual poses relative to the current pose are emphasized instead of the absolute position of each pose. In the case of local motion with a small movement range, changes between the current pose and the others will be magnified, contributing to a more accurate prediction result. 

After the positional and temporal enhanced representations are yielded, the postural information in these representations should be extracted independently in each body group in Feature Fusion Network. Subsequently, prior knowledge about human body structure needs to be transferred between groups without disturbing the previous process. To this end, we propose a multi-stage optimization method for the whole framework. This method sustains the independence of the preceding two processes by optimizing them in sequence.

In summary, the contributions of this paper are threefold:
\begin{itemize}
  \item We propose a positional information encoding method to yield a prediction result that is more robust to global motion.
  \item We propose a temporal information encoding method to yield a prediction result that is more accurate on local motion with a small movement range.
  \item To better utilize the positional and temporal information within each human body group, we introduce a multi-stage optimization method to the whole framework.
\end{itemize}

\begin{figure*}
  \includegraphics[width=\textwidth]{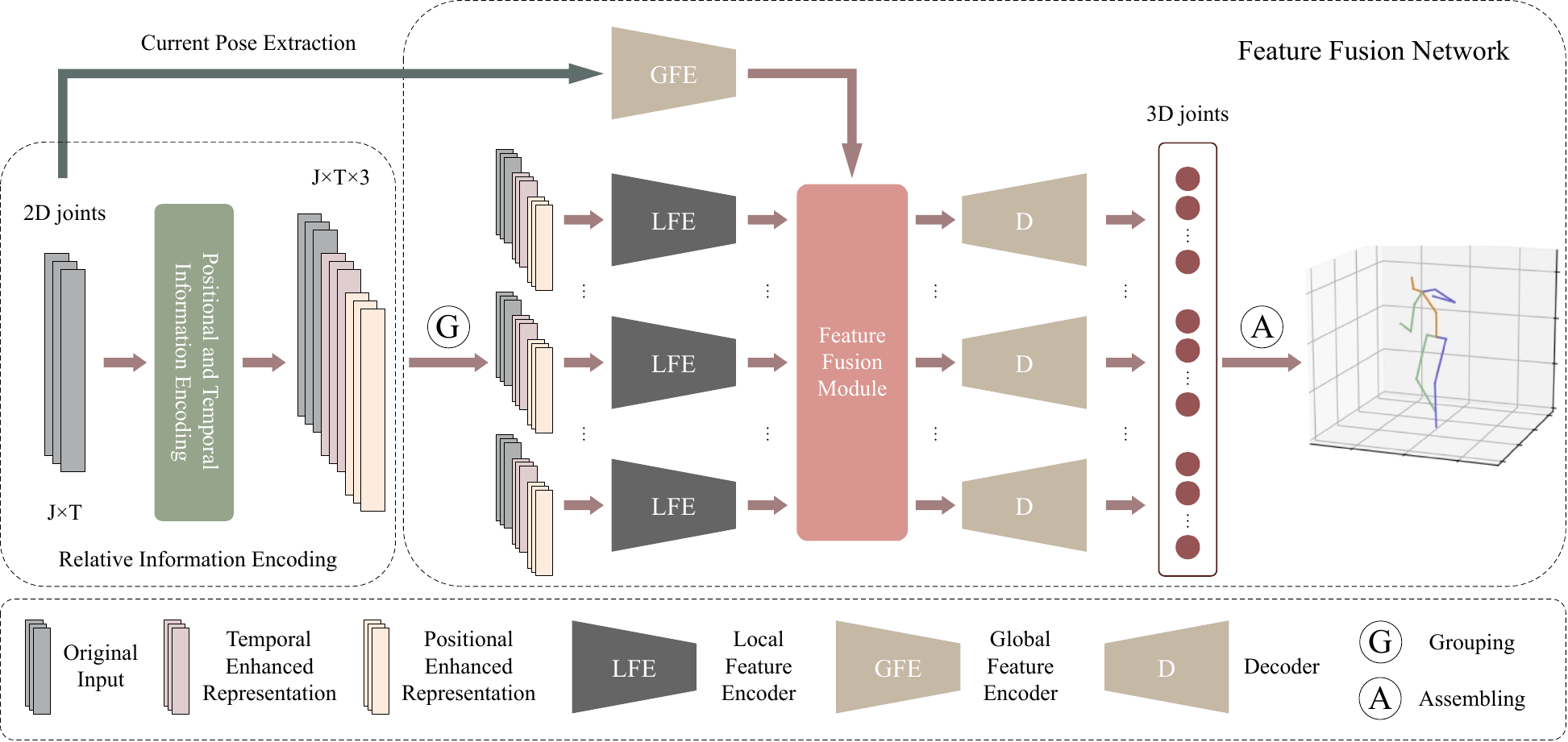}
  \vspace{-0.4cm}
  \caption{An overview of our framework. The positional and temporal information is enhanced by the proposed relative information encoding method. Then, the enhanced representations, together with the original input, are fed into the Feature Fusion Network in which human joints are partitioned into $N$ groups. The local features within a group are captured by the local feature encoder. Additionally, the global features are extracted from the current pose. All features are fused by the Feature Fusion Module and sent to the decoders to yield the final 3D pose. }
  \vspace{-0.2cm}
  \label{img2}
\end{figure*}

\section{Related Works}
There are two categories of methods for 3D human pose estimation. Most of the early works \cite{pavlakos2017coarse,li20143d,tekin2016structured,rogez2017lcr}, which directly regress 3D human joints from original images or videos, are known as one-step methods. Recent approaches \cite{jllo20193d,martinez2017simple,cai2019exploiting,chen20173d,iqbal2020weakly}, which firstly detect 2D keypoints and then infer the corresponding 3D poses, are known as two-step methods. We briefly review the learning-based two-step approaches in this section.
\subsection{Anatomical Grouping Strategy}
Many methods \cite{park20183d,cai2019exploiting,tekin2016structured,ci2019optimizing,cheng2019occlusion,wang2019not,moreno20173d,wang2018drpose3d,wang2011learning} leverage the internal structural constraints that exist between human joints to improve the performance of 3D pose estimation. Recently, the anatomical grouping strategy, which partitions the human body into local groups to capture the commonality of related joints, is widely exploited. Park \textit{et al.} \cite{park20183d} capture the inter-dependencies among joints within a group via a hierarchical relational network. Fang \textit{et al.} \cite{fang2018learning} develop a pose grammar to encode high-order relations among human body parts. Wang \textit{et al.} \cite{wang2019not} divide a human body into different parts with varying levels of degrees of freedom and explicitly model the bi-directional dependencies among parts. Zheng \textit{et al.} \cite{zheng2020joint} regard each joint as an independent group, in which the local features are extracted. Zeng \textit{et al.} \cite{zeng2020srnet} propose a split-and-recombine approach to incorporate low-dimensional global context into the local group. Similarly, we categorize the body parts into five groups and build a baseline called Feature Fusion Network. We come up with a Feature Fusion Module (FFM) to explicitly model the associations between these groups. In contrast to \cite{zeng2020srnet}, which exchanges information in every convolution layer, our approach implements feature fusion only after the encoder networks. In this way, the processes of local feature extraction and feature fusion are separated. Through the proposed multi-stage optimization method, the encoder in each group can work independently without interference from other groups.

\subsection{Positional Information}
Many recent works \cite{mehta2017monocular,sun2017compositional,li2020cascaded,xu2020deep,zhou2017towards,dabral2018learning} propose novel reparameterization methods for 3D human poses at the output end. Sun \textit{et al.} \cite{sun2017compositional} use bones instead of joints to represent human poses, and exploit the joint connection structure to define a compositional loss function for training. Mehta \textit{et al.} \cite{mehta2017monocular} represent joint positions in various ways, such as positions relative to the root, or to the first order parent. Other works \cite{li2020cascaded,xu2020deep} decompose the coordinates of human joints into axis and angle in the polar coordinate system relative to the parent joints, making it easier to yield a unified bone length for a particular person. However, all of the existing approaches only focus on searching for better 3D representations, which capture the posture of the human body. Few efforts have been made to ensure the uniformity of the input and output distribution. As a result, global motion, such as the offset of the camera position, will cause performance degradation. In contrast, we propose a positional information encoding method that represents poses in the relative coordinate system with respect to the root joint both on the input and output side, eliminating the negative effects of the global motion.

\subsection{Temporal Information}
In order to maintain temporal consistency and mitigate jitter in videos, many approaches \cite{cai2019exploiting,xu2020deep,hossain2018exploiting,lin2017recurrent,lin2019trajectory,jllo20193d,liu2020attention,wang2020motion,mehta2017vnect} have been proposed to exploit temporal information for 3D pose inferring in video sequences. Recently, LSTM models \cite{hochreiter1997long} are introduced to 3D human pose estimation. Hossain \textit{et al.} \cite{hossain2018exploiting} propose a sequence-to-sequence model composed of LSTM unit to estimate the 3D pose sequences. Xu \textit{et al.} \cite{xu2020deep} propose to refine the 3D trajectories using a bi-directional LSTM network. Lin \textit{et al.} \cite{lin2019trajectory} process all input frames concurrently, and utilize matrix factorization to deal with the drift problem of LSTM models. Besides, Pavllo \textit{et al.} \cite{jllo20193d} propose a Temporal Convolutional Network (TCN) that performs 1D convolutions over time sequences. A sequence of 2D poses is utilized to jointly estimate the current 3D pose. Compared with the methods using LSTMs, TCN can achieve higher accuracy with fewer parameters. However, all of the 2D poses are treated equally by the temporal 1D convolution used by TCN. The discrepancy between the current pose and other poses is overlooked. Although Liu \textit{et al.} \cite{liu2020attention} apply the attention mechanism to TCN, only tensors in the middle layers are weighted, and the importance of the current pose is still not emphasized explicitly. The relationships of the current pose and the others are implicitly modeled, which increases the burden on the network. As a result, small differences between poses are less likely to be captured by the network, resulting in bad prediction results. Unlike these works, we explicitly encode the temporal information by building a direct correlation between the current frame and the others. This method boosts the estimation performance on local motion with a small movement range.

\section{Proposed Method}
An overview of our framework is illustrated in Figure~\ref{img2}. It takes a sequence of 2D keypoints as input and yields a single 3D pose in the current frame as output. To take advantage of the structural information of the human body, we build a baseline called Feature Fusion Network based on an auto-encoder paradigm. We assign spatially associated joints into the same group, in which shared features can be learned to model the commonality of the grouped joints. Given the fact that predicting joint positions in each group independently may produce physically impossible body movements, we propose a Feature Fusion Module (FFM), similar to \cite{zeng2020srnet}, to depict the dependencies between groups. This module is beneficial for the establishment of structural associations between different groups, and ensure that the generated 3D poses are reasonable. To mitigate the problem of lacking robustness to global motion and inaccurate prediction results of local motion, we propose a relative information encoding method. We utilize this method to yield positional and temporal enhanced representations $(\mathcal{K}_{P},\mathcal{K}_{T})$, which serve as additional inputs to the baseline.

\begin{figure}
  \centering
  \includegraphics[width=\linewidth]{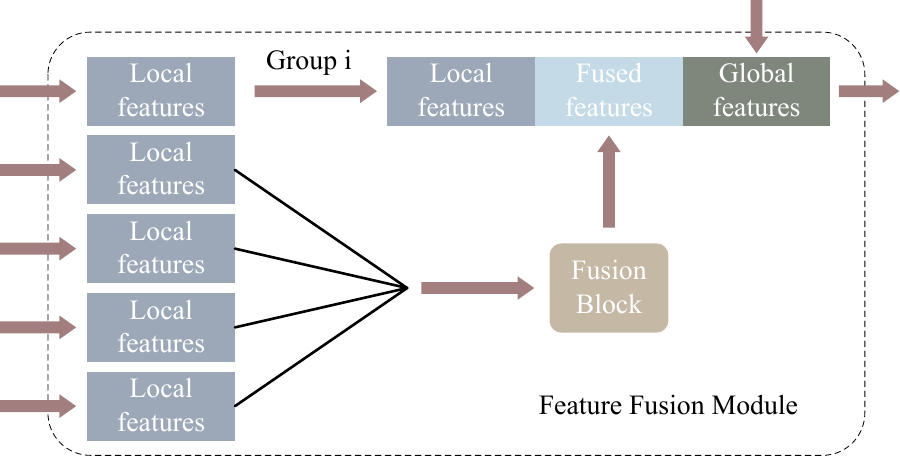}
  \vspace{-0.4cm}
  \caption{Illustration of the Feature Fusion Module. Local features in the current group, fused features produced by the fusion block, and global features produced by the global feature encoder are concatenated and fed into the decoder.}
  \vspace{-0.3cm}
  \label{img3}
\end{figure}

\subsection{Baseline}\label{baseline}

To incorporate the relative information encoding into 3D human pose estimation, we propose a Feature Fusion Network as the baseline in view of the physiological structure of the human body. For each pose, we adopt the grouping strategy used in \cite{park20183d}. We denote a sequence of 2D vectorized poses in image space as $\mathcal{K}=\left\{\mathbf{k}^{j}_{t}\right\}$ , $\mathbf{k}^{j}_{t}={(\mathbf{x}^{j}_{t},\mathbf{y}^{j}_{t})} \in \mathbb{R}^{2}$, and the corresponding 3D poses in the absolute coordinate system as $\mathcal{P}=\left\{\mathbf{p}^{j}_{t}\right\}$ , $\mathbf{p}^{j}_{t}={(\mathbf{X}^{j}_{t},\mathbf{Y}^{j}_{t},\mathbf{Z}^{j}_{t})} \in \mathbb{R}^{3}$, where $j=1,2,\ldots,J, t=1,2,\ldots,T$. ${J}$ is the number of human joints, and ${T}$ is the length of the sequence. For each pose, all human joints are divided into $N=5$ groups, including torso, left arm, right arm, left leg, and right leg. The 2D human joints in each group can be expressed as $\mathcal{K}^{i}=\left\{\mathbf{k}^{j}_{t}\right\}_{j=1}^{J_i}$, where ${J_i}$ is the number of joints in group $i$. 

Local features are extracted from grouped 2D keypoints by the local feature encoder, which uses TCN \cite{jllo20193d} as the backbone. This process can be expressed as:
\begin{equation}
\mathbf{F}_{l}^{i}={\mathcal{E}_{l}^{i}}(\mathcal{K}^{i}, \theta^{i})\label{eq}
\end{equation}
where ${\mathcal{E}_{l}^{i}}(\cdot, \theta)$ stands for the local feature encoder in group $i$. With the intention of enabling the framework to comprehend the global coherence of all human joints that is not interfered by other frames, the global features are extracted from the current pose $\mathcal{K}_{c}=\left\{\mathbf{k}^{j}_{\frac{T}{2}}\right\}_{j=1}^{J}$. This process can be expressed as:
\begin{equation}
\mathbf{F}_{g}={\mathcal{E}_{g}}(\mathcal{K}_{c}, \theta)
\end{equation}
where ${\mathcal{E}_{g}}(\cdot, \theta)$ stands for the global feature encoder.

Although the spatially meaningful patterns are preserved in each group, the connection between different groups is excluded. When inferring 3D poses, the joint positions of other groups are completely unknown to the current group, which is not good for maintaining the consistency of the overall body posture. Ideally, the networks are supposed to account for the continuity of the joints between groups. This motivates us to propose a Feature Fusion Module (FFM) that transfers the information of other groups to the current group. This module can work effectively in combination with the proposed multi-stage optimization method, which will be discussed in Section~\ref{MS}. Figure~\ref{img3} illustrates the structure of the FFM. The fused features are obtained by a fusion block:
\begin{equation}
\mathbf{F}_{f}^{i}={\mathcal{G}_{f}}(\mathcal{M}_{i}, \theta)
\end{equation}
where $\mathcal{M}_{i} = \left\{\mathbf{F}_{l}^{n}|n=1,2, \ldots, N, n \neq i\right\}$, $N$ is the number of groups, and ${\mathcal{G}_{f}}(\cdot, \theta)$ stands for the fusion block. Subsequently, all of the features are concatenated and sent to the decoder to estimate the 3D pose:
\begin{equation}
\tilde{\mathcal{P}}^{i}={\mathcal{D}^{i}}(\mathbf{F}_{l}^{i}\oplus\mathbf{F}_{f}^{i}\oplus\mathbf{F}_{g}, \theta)
\end{equation}
where $\oplus$ is the concatenation operator, and ${\mathcal{D}}(\cdot, \theta)$ stands for the decoder. It should be noticed that the global feature encoder, decoders, and fusion block in the FFM share the same network structure that is proposed by \cite{martinez2017simple}. It consists of two fully connected layers as well as a skip connection.

\begin{figure}
  \centering
  \includegraphics[width=\linewidth]{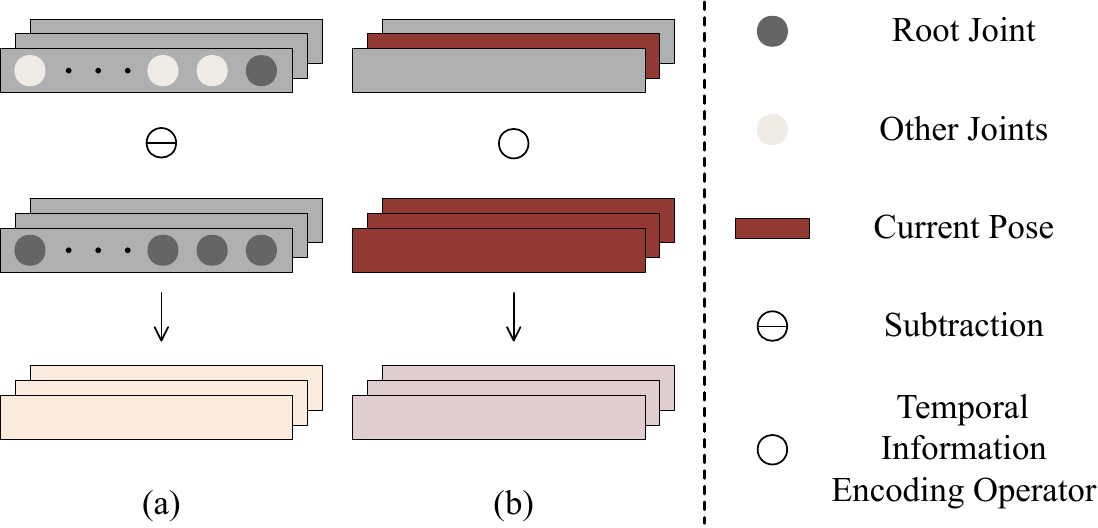}
  \vspace{-0.4cm}
  \caption{Illustration of the relative information encoding method, including (a) positional information encoding and (b) temporal information encoding. }
  \vspace{-0.3cm}
  \label{img11}
\end{figure}

\subsection{Positional Information Encoding}\label{P-ENH}
To yield 3D poses that are less sensitive to global motion, we propose a positional information encoding method. Most of the previous approaches \cite{liu2020attention,martinez2017simple,jllo20193d} estimate 3D joint positions with respect to the root joint. They eliminate the 3D global trajectory for restricting the networks to focus on estimating the posture of the human body. However, none of them consider the uniformity of the input and output forms. At a certain time ${t}$, the goal of the previous networks can be formulated as a function:

\begin{equation}\label{eq1}
\mathcal{F}:\left\{(\mathbf{x}^{j},\mathbf{y}^{j})\right\}\rightarrow \left\{(\mathbf{X}^{j}-\mathbf{X}^{0},\mathbf{Y}^{j}-\mathbf{Y}^{0},\mathbf{Z}^{j}-\mathbf{Z}^{0})\right\}
\end{equation}
where $j=1,2,\ldots,J$. ${(\mathbf{X}^{0},\mathbf{Y}^{0},\mathbf{Z}^{0})}$ represent the 3D coordinates of the root joint (i.e. pelvis). ${t}$ is omitted for simplicity. There is inconsistency in the distribution of the input and output, which will deteriorate the robustness to global motion. For example, when a hand-held camera is capturing a person with the same posture, the camera is prone to be shifted. The overall 2D positions of the person in the videos taken by the camera, before and after the shift, are different, while the relative 2D positions of the corresponding joints stay the same. The overall shift of the 2D coordinates can be regarded as the global motion in this situation, thus equation~(\ref{eq1}) can be reformulated as:
\begin{equation}
\mathcal{F}_{s}:\mathcal{K}_{s}\rightarrow \mathcal{P}_{s}
\end{equation}
where
\begin{equation}\label{eq2}
\mathcal{K}_{s}=\left\{(\mathbf{x}^{j}+\Delta\mathbf{x},\mathbf{y}^{j}+\Delta\mathbf{y})\right\}
\end{equation}
\begin{equation}
\begin{aligned}
\mathcal{P}_{s}=\left\{\right.(&\mathbf{X}^{j}+\Delta\mathbf{X}-(\mathbf{X}^{0}+\Delta\mathbf{X}),\\
&\mathbf{Y}^{j}+\Delta\mathbf{Y}-(\mathbf{Y}^{0}+\Delta\mathbf{Y}),\\
&\mathbf{Z}^{j}+\Delta\mathbf{Z}-(\mathbf{Z}^{0}+\Delta\mathbf{Z}))\left.\right\}\\
=\left\{\right.(&\mathbf{X}^{j}-\mathbf{X}^{0},\mathbf{Y}^{j}-\mathbf{Y}^{0},\mathbf{Z}^{j}-\mathbf{Z}^{0})\left.\right\}
\end{aligned}
\end{equation}
The global motion affects the input, but the output is still the same. In addition to inferring the 3D posture of the human body, the network is supposed to learn a mapping function from multiple inputs to the same output. Namely, the network needs to complete the conversion from the absolute 2D coordinates in image space to the relative coordinates in 3D space. When the network 
fails to learn this mapping relationship, a person with the same posture shot by the camera before and after the shift will correspond to different 3D pose estimation results, which is undesirable.

\begin{figure}
  \centering
  \includegraphics[width=\linewidth]{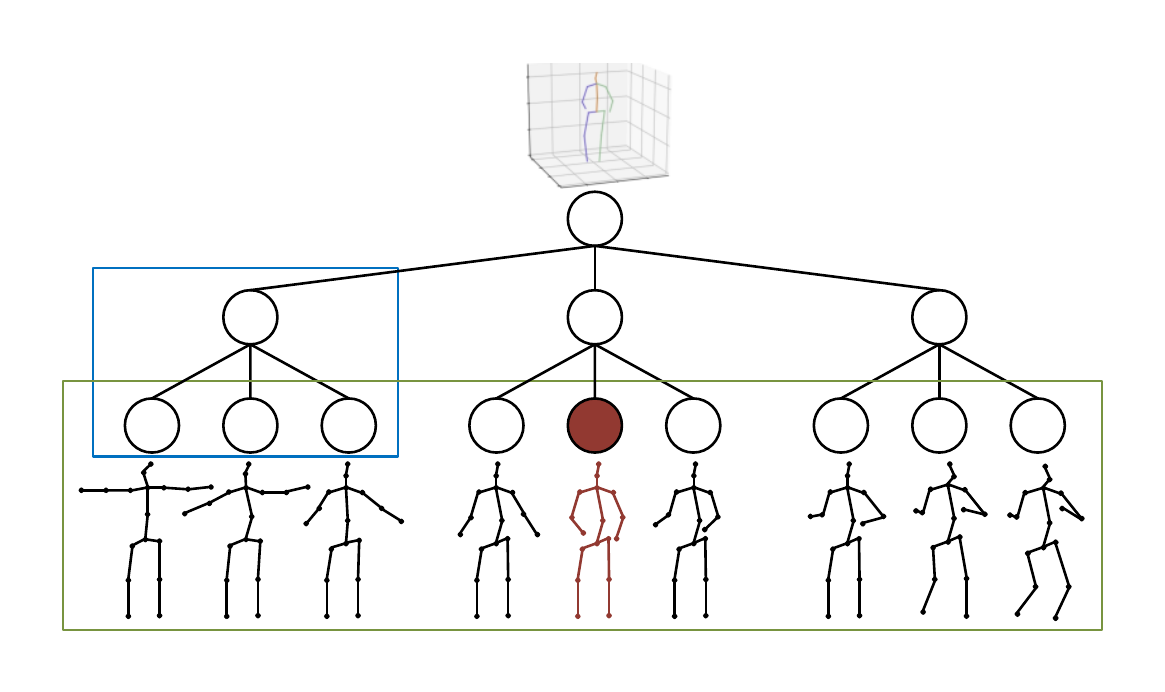}
  \vspace{-0.8cm}
  \caption{Shortcomings of TCN. Green rectangle: TCN treats all poses equally, not knowing which one needs to be lifted to 3D space. Blue rectangle: the convolution operation in shallow layers only aggregates information in a local region that does not include the current pose (red).}
  \vspace{-0.2cm}
  \label{img4}
\end{figure}

To address this issue, we encode the positional information in the input side by subtracting the coordinates of the root joint from the positions of all other joints. As shown in Figure~\ref{img11}(a), the positional information encoding can be expressed as:
\begin{equation}
\mathcal{K}_{P}=\left\{{\mathbf{k}}^{j}-{\mathbf{k}}^{0}\right\}_{j=1}^{J}
\end{equation}
$t$ is omitted for simplicity. Then the goal of the network can be rewritten as:
\begin{equation}\label{eq3}
\mathcal{F}_{P}:\left\{(\mathbf{x}^{j}-\mathbf{x}^{0},\mathbf{y}^{j}-\mathbf{y}^{0})\right\}\rightarrow \left\{(\mathbf{X}^{j}-\mathbf{X}^{0},\mathbf{Y}^{j}-\mathbf{Y}^{0},\mathbf{Z}^{j}-\mathbf{Z}^{0})\right\}
\end{equation}
where ${(\mathbf{x}^{0},\mathbf{y}^{0})}$ represent the 2D coordinates of the root joint. When the global motion occurs, equation~(\ref{eq3}) can be reformulated as:
\begin{equation}
\mathcal{F}_{Ps}:\mathcal{K}_{Ps}\rightarrow \mathcal{P}_{s}
\end{equation}
where
\begin{equation}
\begin{aligned}
\mathcal{K}_{Ps}=\left\{\right.(&\mathbf{x}^{j}+\Delta\mathbf{x}-(\mathbf{x}^{0}+\Delta\mathbf{x}),\\
&\mathbf{y}^{j}+\Delta\mathbf{y}-(\mathbf{y}^{0}+\Delta\mathbf{y}))\left.\right\}\\
=\left\{\right.(&\mathbf{x}^{j}-\mathbf{x}^{0},\mathbf{y}^{j}-\mathbf{y}^{0})\left.\right\}
\end{aligned}
\end{equation}
and $\mathcal{P}_{s}$ remains the same. Hence, $\mathcal{F}_{Ps}=\mathcal{F}_{P}$. In this way, the same posture with different absolute 2D coordinates will correspond to a common positional enhanced representation, which reduces the difficulty for the network to yield the same prediction result. The positional information encoding forces the network to only capture meaningful information related to the human posture instead of 2D global trajectory. This is conducive for the network to become more robust to global motion. Besides, the 2D global trajectory cannot be discarded as it is beneficial to 3D pose inferring. Its necessity is assessed in the supplementary material.

\subsection{Temporal Information Encoding}\label{T-ENH}
In addition to enhancing the positional information, we propose a temporal information encoding method to alleviate the issue of bad performance on local motion with a small movement range. Previous methods \cite{jllo20193d,liu2020attention} use Temporal Convolutional Network (TCN) that takes a sequence of 2D poses as input and produces a single 3D pose. Figure~\ref{img4} shows two shortcomings of TCN: 1) Although several 2D poses within a short period of time are fed into the TCN, only a single pose in the middle of the sequence is supposed to be estimated. All other poses play an auxiliary role in maintaining the continuity of the current pose in the time domain. Therefore, they are less important than the current pose. TCN treats all poses equally, not knowing which pose needs to be lifted to 3D space. The relative positional relationships between the current pose and the others are not explicitly emphasized. 2) According to the attribute of convolution operation, the shallow layers in neural networks have a small receptive field, and only aggregate information in a local region, while deep layers have a larger receptive field. This indicates that poses far away from the current pose cannot obtain any information related to the current pose till deeper layers are reached. These two shortcomings will cause that the network doesn’t pay enough attention to the changes occurring around the current pose. When the movement range of poses is relatively small, these changes are difficult to be captured by the network, resulting in bad prediction performance on small movements.

As we use TCN as the backbone of the local feature encoders, we propose a temporal information encoding method to overcome these shortcomings. We produce the temporal enhanced representation to model the relationships between the current pose and other poses. This method drives the network to learn the impact of the current pose on other poses. It allows all poses to start focusing on the temporal correlation with the current pose, whether they are far from it or near to it in the time domain, from the shallow layers of the network. In other words, the network concentrates on the position changes around the current pose rather than the absolute position of each pose. When the local motion with a small movement range occurs, these changes will be magnified, which is helpful for fine-grained modeling of 3D poses. As shown in Figure~\ref{img11}(b), the temporal information encoding can be formulated as follows. 
\begin{equation}
\mathcal{K}_{T}=\left\{{\mathbf{k}}_{t}\circ{\mathbf{k}}_{\frac{T}{2}}\right\}_{t=1}^{T}
\end{equation}
where $\circ$ can be any vector operator, such as inner-product, cross-product, cosine similarity, and subtraction. $j$ is omitted for simplicity. Experiments are conducted to evaluate the effectiveness of different operators in the supplementary material.

The positional and temporal enhanced representations produced by the proposed relative information encoding method are utilized as additional inputs to the Feature Fusion Network. Thus, equation~(\ref{eq}) can be reformulated as:

\begin{equation}
\mathbf{F}_{l}^{i}={\mathcal{E}_{l}^{i}}(\mathcal{K}^{i},\mathcal{K}_{P}^{i},\mathcal{K}_{T}^{i}, \theta^{i})
\end{equation}

\begin{figure}
  \centering
  \includegraphics[width=\linewidth]{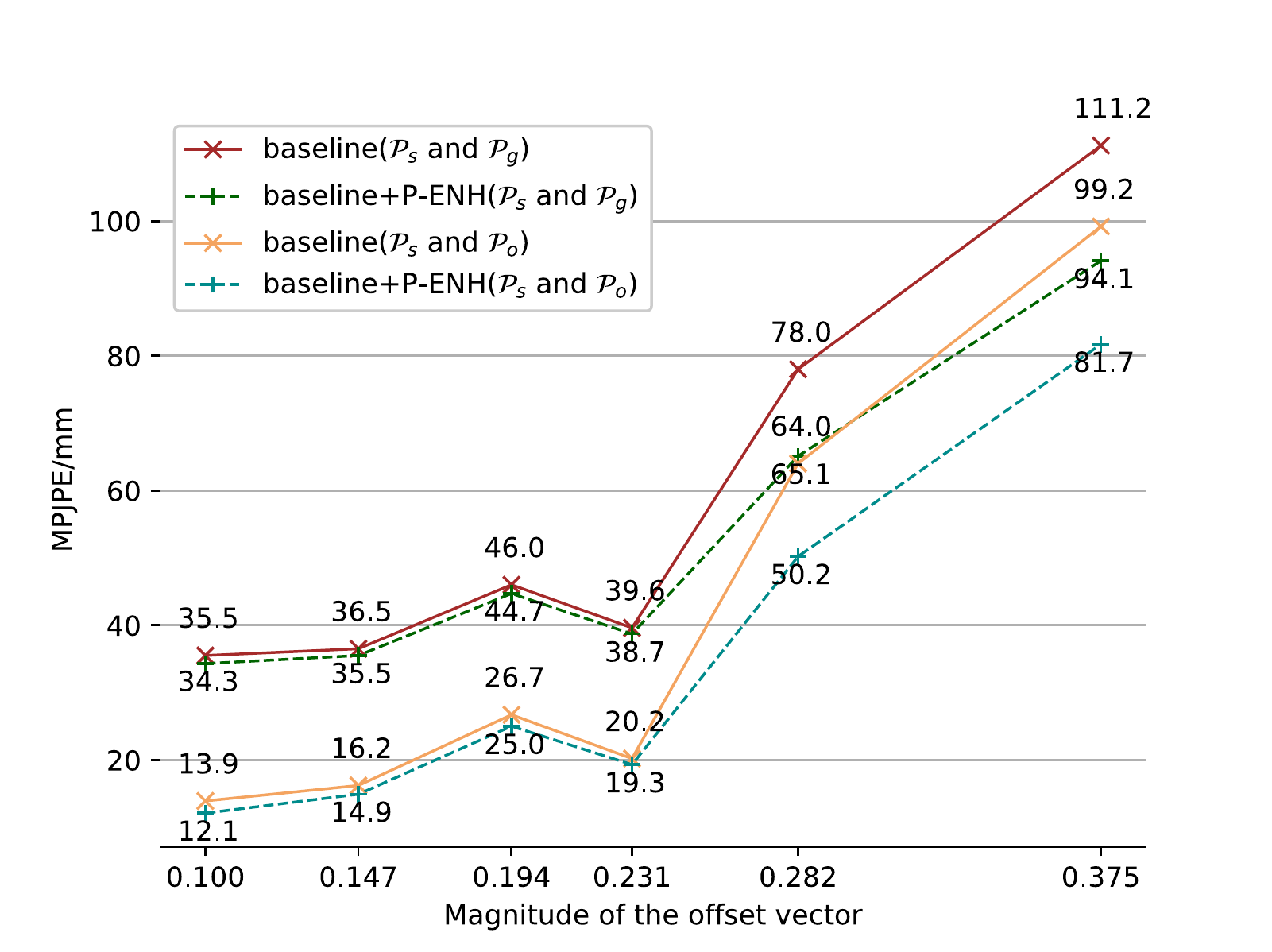}
  \vspace{-0.5cm}
  \caption{Testing error of $\mathcal{P}_{s}$ (red and green) as well as the similarity measurement between $\mathcal{P}_{s}$ and $\mathcal{P}_{o}$ (yellow and cyan) in the case of global motion caused by simulated camera movement.}
  \vspace{-0.3cm}
  \label{img6}
\end{figure}

\subsection{Multi-Stage Optimization}\label{MS}
We propose a multi-stage optimization method to fully exploit the positional and temporal information in each group for the following consideration. The information exchange should be implemented between groups selectively and comprehendingly under the condition of preserving the independence of the encoding process. In \cite{zeng2020srnet}, the processes of capturing the internal commonality in each group and exchanging useful information between groups are blended together. This will cause that the features in each group are interfered with by other groups. Therefore, we separate the preceding two processes in the Feature Fusion Network and apply the proposed multi-stage optimization method to our framework. We retain the independence of the local feature encoding and feature fusion through this method, thus clarifying the role of these two processes. 

Feature Fusion Network can be divided into three parts: encoders, decoders, and FFM. In the first stage of optimization, We temporarily delete the FFM so that the local feature encoders can extract regional features of related joints independently. At the same time, the global feature encoder accounts for the overall coherence of the human body. The local and global features are concatenated and sent to the decoders, which ensure the integrity of the framework for end-to-end training. The parameters of the local and global feature encoders are preserved for the second stage of optimization. In the second stage, only the FFM and decoders participate in optimization. The parameters of the encoders are loaded and fixed, while the parameters of the decoders in the first stage are abandoned. The decoders are retrained in this stage. In this way, the FFM sufficiently comprehends the inter-group dependencies without disturbing the optimization process of the encoders. In the third stage, the entire framework is finetuned to make the encoders work with the decoders and FFM more effectively. With the help of the proposed multi-stage optimization method, posture-related information in positional and temporal representations can be extracted independently in each group, and positive interaction is formed between groups to facilitate 3D human pose estimation.

\begin{figure}
  \centering
  \includegraphics[width=\linewidth]{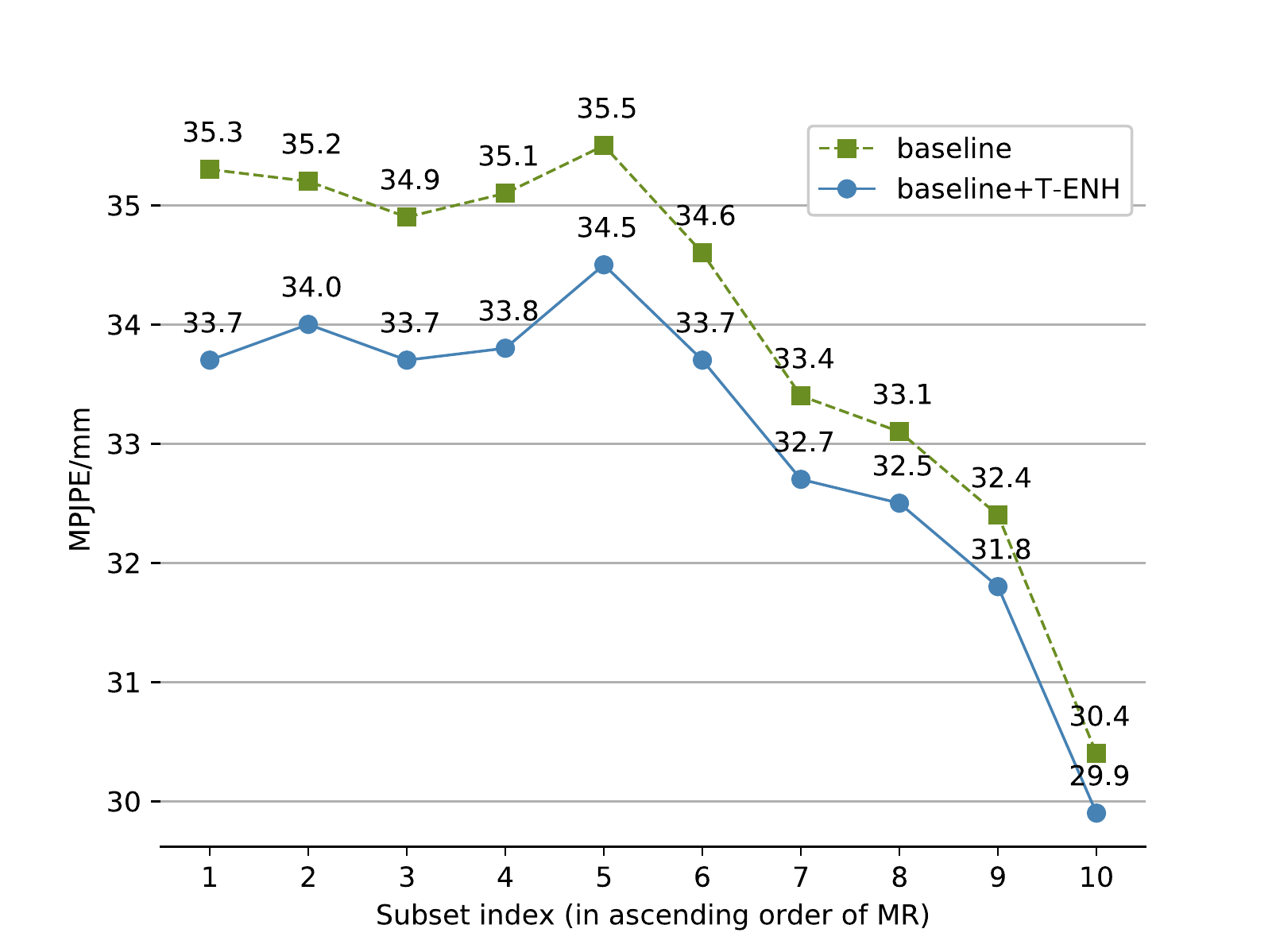}
  \vspace{-0.5cm}
  \caption{Testing error of the baseline and the baseline+T-ENH in ten subsets with different movement ranges.}
  \label{img7}
\end{figure}

\begin{table}
  \centering
  \caption{Ablation analysis of the proposed framework when different components are applied. P-ENH and T-ENH refer to positional and temporal enhanced representations respectively. M-S OPT: multi-stage optimization.}
    \begin{tabular}{ccc}
    \toprule
    Method & MPJPE(mm)  & $\Delta$ \\
    \midrule
    Baseline & 34.1&-  \\
    +P-ENH & 33.4&0.7  \\
    +T-ENH & 33.1&1.0  \\
    +P-ENH+T-ENH &32.3&1.8 \\
    +P-ENH+T-ENH(M-S OPT)&30.1 & 4.0\\
    \bottomrule
    \end{tabular}
  \label{tab4}
  \vspace{-0.3cm}
\end{table}

\begin{table*}[ht]\small
  \centering
  \renewcommand\tabcolsep{4.0pt}
  \caption{Results on Human3.6M in millimeter under \textit{Protocol\#1} (MPJPE). Top table: 2D poses obtained by CPN are used as inputs. Bottom table: the ground truth of 2D poses are used as inputs. The best result is shown in bold, and the second-best result is underlined.}
    \begin{tabular}{c|ccccccccccccccc|c}
    \toprule
    Method & Dir. & Disc. & Eat & Greet & Phone & Photo & Pose & Pur. & Sit & SitD. & Smoke & Wait & WalkD. & Walk & WalkT. & Avg \\
    \midrule
    Martinez \textit{et al.} \cite{martinez2017simple} ICCV'17 & 51.8&56.2&58.1&59.0&69.5&78.4&55.2&58.1&74.0&94.6&62.3&59.1&65.1&49.5&52.4&62.9 \\
    Pavlakos \textit{et al.} \cite{pavlakos2017coarse} CVPR'17& 48.5&54.4&54.4&52.0&59.4&65.3&49.9&52.9&65.8&71.1&56.6&52.9&60.9&44.7&47.8&56.2 \\
    Fang \textit{et al.} \cite{fang2018learning} AAAI'18& 50.1&54.3&57.0&57.1&66.6&73.3&53.4&55.7&72.8&88.6&60.3&57.7&62.7&47.5&50.6&60.4 \\
    Hossain \textit{et al.} \cite{hossain2018exploiting} ECCV'18& 48.4&50.7&57.2&55.2&63.1&72.6&53.0&51.7&66.1&80.9&59.0&57.3&62.4&46.6&49.6&58.3 \\
    Lee \textit{et al.} \cite{lee2018propagating} ECCV'18& \underline{40.2}&49.2&47.8&52.6&50.1&75.0&50.2&43.0&55.8&73.9&54.1&55.6&58.2&43.3&43.3&52.8 \\
    Pavllo \textit{et al.} \cite{jllo20193d} CVPR'19& 45.2&46.7&43.3&45.6&48.1&55.1&44.6&44.3&57.3&65.8&47.1&44.0&49.0&32.8&33.9&46.8 \\
    Lin \textit{et al.} \cite{lin2019trajectory} BMVC'19& 42.5&44.8&42.6&44.2&48.5&57.1&42.6&\underline{41.4}&56.5&64.5&47.4&43.0&48.1&33.0&35.1&46.6 \\
    Xu \textit{et al.} \cite{xu2020deep} CVPR'20
    &\textbf{37.4}&\textbf{43.5}&42.7&\underline{42.7}&46.6&59.7&\textbf{41.3}&45.1&\textbf{52.7}&\textbf{60.2}&45.8&43.1&47.7&33.7&37.1&45.6\\
    Liu \textit{et al.} \cite{liu2020attention} CVPR'20& 41.8&44.8&\textbf{41.1}&44.9&47.4&\underline{54.1}&43.4&42.2&56.2&63.6&\underline{45.3}&43.5&45.3&\underline{31.3}&\underline{32.2}&45.1 \\
    Zeng \textit{et al.} \cite{zeng2020srnet} ECCV'20& 46.6&47.1&43.9&\textbf{41.6}&\textbf{45.8}&\textbf{49.6}&46.5&\textbf{40.0}&\underline{53.4}&61.1&46.1&\underline{42.6}&\textbf{43.1}&31.5&32.6&\underline{44.8} \\

    \midrule
    Ours ($T=243$ CPN)  &40.8&\underline{44.5}&\underline{41.4}&\underline{42.7}&\underline{46.3}&55.6&\underline{41.8}&41.9&53.7&\underline{60.8}&\textbf{45.0}&\textbf{41.5}&\underline{44.8}&\textbf{30.8}&\textbf{31.9}&\textbf{44.3}\\
    \midrule
    \midrule
    Martinez\textit{et al.} \cite{martinez2017simple} ICCV'17 &37.7&44.4&40.3&42.1&48.2&54.9&44.4&42.1&54.6&58.0&45.1&46.4&47.6&36.4&40.4&45.5 \\
    Hossain \textit{et al.} \cite{hossain2018exploiting} ECCV'18 &35.2&40.8&37.2&37.4&43.2&44.0&38.9&35.6&42.3&44.6&39.7&39.7&40.2&32.8&35.5&39.2  \\
    Lee \textit{et al.} \cite{lee2018propagating} ECCV'18 &\underline{32.1}&36.6&34.4&37.8&44.5&49.9&40.9&36.2&44.1&45.6&35.3&35.9&37.6&30.3&35.5&38.4  \\
    Pavllo \textit{et al.} \cite{jllo20193d} CVPR'19 &35.2&40.2&32.7&35.7&38.2&45.5&40.6&36.1&48.8&47.3&37.8&39.7&38.7&27.8&29.5&37.8  \\
    Lin \textit{et al.} \cite{lin2019trajectory} BMVC'19& -&-&-&-&-&-&-&-&-&-&-&-&-&-&-&32.8 \\
    Liu \textit{et al.} \cite{liu2020attention} CVPR'20 &34.5&37.1&33.6&34.2&32.9&37.1&39.6&35.8&40.7&41.4&33.0&33.8&33.0&26.6&26.9&34.7   \\
    Zeng \textit{et al.} \cite{zeng2020srnet} ECCV'20 &34.8&\underline{32.1}&\textbf{28.5}&\underline{30.7}&\underline{31.4}&\underline{36.9}&\underline{35.6}&\underline{30.5}&\underline{38.9}&\underline{40.5}&\underline{32.5}&\underline{31.0}&\underline{29.9}&\textbf{22.5}&\textbf{24.5}&\underline{32.0}   \\
    \midrule
    Ours ($T=243$ GT)  & \textbf{29.5}&\textbf{30.8}&\underline{28.8}&\textbf{29.1}&\textbf{30.7}&\textbf{35.2}&\textbf{31.7}&\textbf{27.8}&\textbf{34.5}&\textbf{36.0}&\textbf{30.3}&\textbf{29.4}&\textbf{28.9}&\underline{24.1}&\underline{24.7}&\textbf{30.1}\\
    \bottomrule
    \end{tabular}
  \label{tab1}
\end{table*}

\begin{table*}[ht]\small
  \centering
  \renewcommand\tabcolsep{4.0pt}
  \caption{Results on Human3.6M after rigid alignment in millimeter under \textit{Protocol\#2} (P-MPJPE).  }
    \begin{tabular}{c|ccccccccccccccc|c}
    \toprule
    Method & Dir. & Disc. & Eat & Greet & Phone & Photo & Pose & Pur. & Sit & SitD. & Smoke & Wait & WalkD. & Walk & WalkT. & Avg \\
    \midrule
    Martinez \textit{et al.} \cite{martinez2017simple} ICCV'17& 39.5&43.2&46.4&47.0&51.0&56.0&41.4&40.6&56.5&69.4&49.2&45.0&49.5&38.0&43.1&47.7 \\
    Pavlakos \textit{et al.} \cite{pavlakos2017coarse} CVPR'17& 34.7&39.8&41.8&38.6&42.5&47.5&38.0&36.6&50.7&56.8&42.6&39.6&43.9&32.1&36.5&41.8 \\
    Fang \textit{et al.} \cite{fang2018learning} AAAI'18& 38.2&41.7&43.7&44.9&48.5&55.3&40.2&38.2&54.5&64.4&47.2&44.3&47.3&36.7&41.7&45.7 \\
    Hossain \textit{et al.} \cite{hossain2018exploiting} ECCV'18& 35.7&39.3&44.6&43.0&47.2&54.0&38.3&37.5&51.6&61.3&46.5&41.4&47.3&34.2&39.4&44.1 \\
    Pavllo \textit{et al.} \cite{jllo20193d} CVPR'19& 34.1&36.1&34.4&37.2&36.4&42.2&34.4&33.6&45.0&52.5&37.4&33.8&37.8&25.6&27.3&36.5 \\
    Xu \textit{et al.} \cite{xu2020deep} CVPR'20
    &\textbf{31.0}&\textbf{34.8}&34.7&\textbf{34.4}&36.2&43.9&\textbf{31.6}&33.5&\textbf{42.3}&\underline{49.0}&37.1&33.0&39.1&26.9&31.9&36.2\\
    Liu \textit{et al.} \cite{liu2020attention} CVPR'20& \underline{32.3}&\underline{35.2}&\underline{33.3}&35.8&\underline{35.9}&\textbf{41.5}&33.2&\underline{32.7}&44.6&50.9&\underline{37.0}&\underline{32.4}&\underline{37.0}&\underline{25.2}&\underline{27.2}&\underline{35.6} \\

    \midrule
    Ours ($T=243$ CPN) & 32.5&36.2&\textbf{33.2}&\underline{35.3}&\textbf{35.6}&\underline{42.1}&\underline{32.6}&\textbf{31.9}&\underline{42.6}&\textbf{47.9}&\textbf{36.6}&\textbf{32.1}&\textbf{34.8}&\textbf{24.2}&\textbf{25.8}&\textbf{35.0} \\
    \bottomrule
    \end{tabular}
    \vspace{-0.3cm}
  \label{tab2}
\end{table*}

\section{Experiments}

\subsection{Datasets and Evaluation Protocols}
We train and evaluate our method on two publicly available datasets: Human3.6M \cite{ionescu2013human3} and HumanEva-I \cite{sigal2010humaneva}. 

Human3.6M is one of the largest indoor datasets. It contains 3.6 million different human poses for 11 subjects from four synchronized cameras, organized into 15 scenarios. Following the previous approaches \cite{martinez2017simple,liu2020attention,jllo20193d,wang2019not}, we use five subjects for training (S1, S5, S6, S7, S8) and two subjects for testing (S9 and S11). 

HumanEva-I is a smaller dataset compared to Human3.6M. It consists of three subjects with six actions. In the same manner as \cite{liu2020attention,pavlakos2017coarse}, we conduct the training and evaluating process by subject on two actions: Walk and Jog.

Two commonly used protocols are utilized for evaluation in our experiments. \textit{Protocol\#1} refers to Mean Per Joint Position Error (MPJPE) that measures the average Euclidean distance between the estimated joint positions and the ground truth positions. \textit{Protocol\#2}, which is denoted as P-MPJPE, refers to the error after the predicted pose is aligned to the ground truth using Procrustes analysis \cite{gower1975generalized}.

\subsection{Ablation Studies}\label{ablation}
In order to verify the effectiveness of the proposed framework, we analyze in detail the role played by each module in this section. The ablation studies are conducted on Human3.6M dataset under \textit{Protocol\#1}. Our framework is trained based on the ground truth of 2D poses. We use subtraction as the temporal information encoding operator.

Table ~\ref{tab4} reports the performance improvement brought by all components. We can observe that the positional and temporal enhanced representations reduce the error by 0.7mm and 1.0mm respectively. These two representations result in a total performance gain of 1.8mm. Besides, it can be seen that the multi-stage optimization method brings performance improvement to the entire framework as it solves the problem of negative interaction between the encoders and the FFM.\\

\textbf{Positional information encoding.} To validate the effectiveness of the positional information encoding, we carry out experiments based on the same posture with different absolute 2D positions. We manually add a random global offset to the input poses to simulate the global motion caused by camera movement. The absolute coordinates of 2D keypoints change, but the relative 2D coordinates with respect to the root joint remain fixed. In this case, multiple 2D inputs correspond to the same 3D pose. 

Since the input coordinates are normalized to lie between -1 and 1, we need to make sure that the shifted input will not go outside this range either. We randomly select an offset vector $(\Delta\mathbf{x},\Delta\mathbf{y})$, where $\Delta\mathbf{x},\Delta\mathbf{y} \in (-a,a)$. $a$ is set to 0.2 empirically. The shifted input can be expressed as equation~(\ref{eq2}), where $\mathbf{x}^{j}+\Delta\mathbf{x},\mathbf{y}^{j}+\Delta\mathbf{y} \in (-1,1)$. We denote different 3D poses as follows.
\begin{itemize}
 \item $\mathcal{P}_{o}$: the 3D pose predicted from the original input.
 \item $\mathcal{P}_{s}$: the 3D pose predicted from the shifted input.
 \item $\mathcal{P}_{g}$: the ground truth of the 3D pose.

\end{itemize}
MPJPE is computed between: $\mathcal{P}_{o}$ and $\mathcal{P}_{g}$, $\mathcal{P}_{s}$ and $\mathcal{P}_{g}$, $\mathcal{P}_{o}$ and $\mathcal{P}_{s}$. The purpose of computing MPJPE between $\mathcal{P}_{s}$ and $\mathcal{P}_{g}$ is to examine the testing error of $\mathcal{P}_{s}$. The purpose of computing MPJPE between $\mathcal{P}_{o}$ and $\mathcal{P}_{s}$ is to examine the consistency of the output results under the circumstance of global motion caused by simulated camera movement. Figure~\ref{img6} shows the results of the baseline and baseline with the positional enhanced representation (baseline+P-ENH) on six randomly selected offset vectors. The magnitude of the offset vector refers to $\sqrt{(\Delta\mathbf{x})^2+(\Delta\mathbf{y})^2}$. The MPJPE between $\mathcal{P}_{o}$ and $\mathcal{P}_{g}$ of the baseline and baseline+P-ENH is 34.1mm and 33.4mm respectively. The red and green lines show the MPJPE between $\mathcal{P}_{s}$ and $\mathcal{P}_{g}$. It can be seen that the baseline+P-ENH is superior to the baseline when the input is shifted. As the magnitude of the offset vector becomes larger, the accuracy of the 3D pose estimation drops significantly and the advantage of baseline+P-ENH is more prominent. The yellow and cyan lines show the MPJPE between $\mathcal{P}_{s}$ and $\mathcal{P}_{o}$. It is shown that the baseline+P-ENH achieves better consistency between $\mathcal{P}_{s}$ and $\mathcal{P}_{o}$ than the baseline. As the magnitude of the offset vector becomes larger, the consistency gap between these two methods is more noticeable. When the magnitude of the offset vector is 0.375, the baseline+P-ENH achieves 17.1mm (about 15.3\%) and 17.5mm (about 17.6\%) improvement in MPJPE between $\mathcal{P}_{s}$ and $\mathcal{P}_{g}$ as well as $\mathcal{P}_{o}$ and $\mathcal{P}_{s}$ respectively against the baseline. In short, the baseline with the positional enhanced representation is more robust to global motion than the baseline. \\


\textbf{Temporal information encoding.} We examine the impact of the temporal information encoding by carrying out experiments on local motion with different movement ranges. As the proposed framework takes a sequence of poses as input, we utilize the Mean Square Error (MSE) between the current pose and other poses in the same sequence to define the Movement Range (MR):
\begin{equation}
\mathbf{MR}=\frac{1}{T}\frac{1}{J}\sum\limits_{t=1}^{T}\sum\limits_{j=1}^{J}\left\|\mathbf{k}_{t}^{j}-\mathbf{k}_{\frac{T}{2}}^{j}\right\|_{2}
\end{equation}
where$\left\|\mathbf{k}_{t}^{j}-\mathbf{k}_{\frac{T}{2}}^{j}\right\|_{2}$ is the Euclidean distance between $\mathbf{k}_{t}^{j}$ and $\mathbf{k}_{\frac{T}{2}}^{j}$. A large mean of the differences between the current pose and the others corresponds to a large MR of this sequence, and vice versa. We divide the poses in the testing set into tens subsets, which are arranged in ascending order according to their MR. Figure~\ref{img7} shows the results of the baseline and baseline with the temporal enhanced representation (baseline+T-ENH) on these subsets. It is shown that poses with a small movement range (left side) are harder to predict than poses with a large movement range (right side). The reason is that there is less information available for the networks to utilize in the time domain. The changes of the absolute 2D coordinates of poses with a large movement range are significant, so the baseline can easily learn the relationships between the current pose and the others. However, the absolute 2D coordinates of poses with a small movement range are basically unchanged within the input sequence, resulting in a lot of redundant information sent to the network. It is difficult for the baseline to concentrate on the subtle changes between different poses in this case. In contrast, the temporal enhanced representation can magnify the differences in poses with a small movement range and improve the performance of the network to a greater extent. Compared to the baseline, the baseline+T-ENH decreases the MPJPE by 1.6mm in the subset with the smallest movement range and 0.5mm in the subset with the largest movement range. This demonstrates the effectiveness of the temporal information encoding on local motion with a small movement range.

\begin{table}\small
  \centering
  \caption{Results on HumanEva-I after rigid alignment in millimeter under \textit{Protocol\#2} (P-MPJPE). (\textsuperscript{$\ast$}) : the high error on "Walk" of S3 is due to corrupted mocap data, thus this value is not involved in the calculation of the average. }
    \begin{tabular}{c|ccc|ccc|c}
    \toprule
   \multirow{2}{*}{Method}   & \multicolumn{3}{c|}{Walk} &\multicolumn{3}{c|}{Jog} &\multirow{2}{*}{Avg} \\

    &S1&S2&S3&S1&S2&S3\\
    \midrule
    
    Martinez \textit{et al.} \cite{martinez2017simple} & 19.7&17.4&46.8&26.9&18.2&18.6&24.6 \\
    Pavlakos \textit{et al.} \cite{pavlakos2017coarse} & 22.3&19.5&\textbf{29.7}&28.9&21.9&23.8&24.4 \\
    Lee \textit{et al.} \cite{lee2018propagating}&18.6&19.9&\underline{30.5}&25.7&16.8&17.7&21.5\\
    Pavllo \textit{et al.} \cite{jllo20193d} &13.9&10.2&46.6\textsuperscript{$\ast$}&20.9&13.1&13.8&14.3 \\
    Liu \textit{et al.} \cite{liu2020attention}&\underline{13.8}&\underline{10.0}&46.4\textsuperscript{$\ast$}&\textbf{20.2}&\underline{12.3}&\underline{12.9}&\underline{13.8}\\

    \midrule
    Ours ($T=27$) &\textbf{13.5}&\textbf{9.4}&47.2\textsuperscript{$\ast$}&\underline{20.3}&\textbf{12.1}&\textbf{12.8}&\textbf{13.6} \\
    \bottomrule
    \end{tabular}
    \vspace{-0.3cm}
  \label{tab3}
\end{table}

\subsection{Comparison with State-of-the-Art Methods}
\textbf{Results on Human3.6M.} Table ~\ref{tab1}-\ref{tab2} show the comparison results of the proposed method and the previous approaches on Human3.6M dataset under two protocols. Our method is compatible with any 2D keypoint detector. Specifically, we use the results of Cascaded Pyramid Network (CPN)  \cite{chen2018cascaded} as the inputs to the proposed framework. Compared to the state-of-the-art method \cite{zeng2020srnet}, our model achieves promising results with 44.3mm in MPJPE and 35.0mm in P-MPJPE. We also train our model on the ground truth of 2D poses. Our model achieves 30.1mm in MPJPE and improves the lower bound of \cite{zeng2020srnet} by about 5.9\%. The qualitative results are provided in the supplementary material.

\textbf{Results on HumanEva-I.} As shown in Table ~\ref{tab3}, we evaluate our method in terms of \textit{Protocol\#2} on HumanEva-I dataset. We use Mask R-CNN \cite{he2017mask} as the 2D detector. Besides, we choose $T=27$ in this experiment. Our model achieves 13.6mm in P-MPJPE, outperforming the previous approaches.

\section{Conclusion}
In this paper, we introduce a relative information encoding method that provides explicit and strong priors of the positional relationships within all human joints as well as the temporal relationships between poses in a sequence. The proposed method facilitates plausible 3D pose estimation on local motion with a small movement range and yields a more robust result to global motion. Additionally, a multi-stage optimization method is proposed for the whole framework. Experiments show that our method outperforms state-of-the-art performance on Human3.6M and HumanEva-I datasets.

\begin{acks}
This work was supported by National Key Research and Development Project, PKU-Baidu Fund (2019BD003) and High-performance Computing Platform of Peking University.
\end{acks}


\bibliographystyle{ACM-Reference-Format}

\onecolumn
\begin{center}
\begin{Huge} 
\textbf{Supplementary Material}
\vspace{0.6cm}
\end{Huge}
\end{center}
\begin{multicols}{2}






\setcounter{equation}{0}
\setcounter{figure}{0}
\setcounter{table}{0}
\setcounter{page}{1}
\setcounter{section}{0}
\makeatletter

\section{Qualitative Results}
\begin{figure*}
  \centering
  \includegraphics[width=\textwidth]{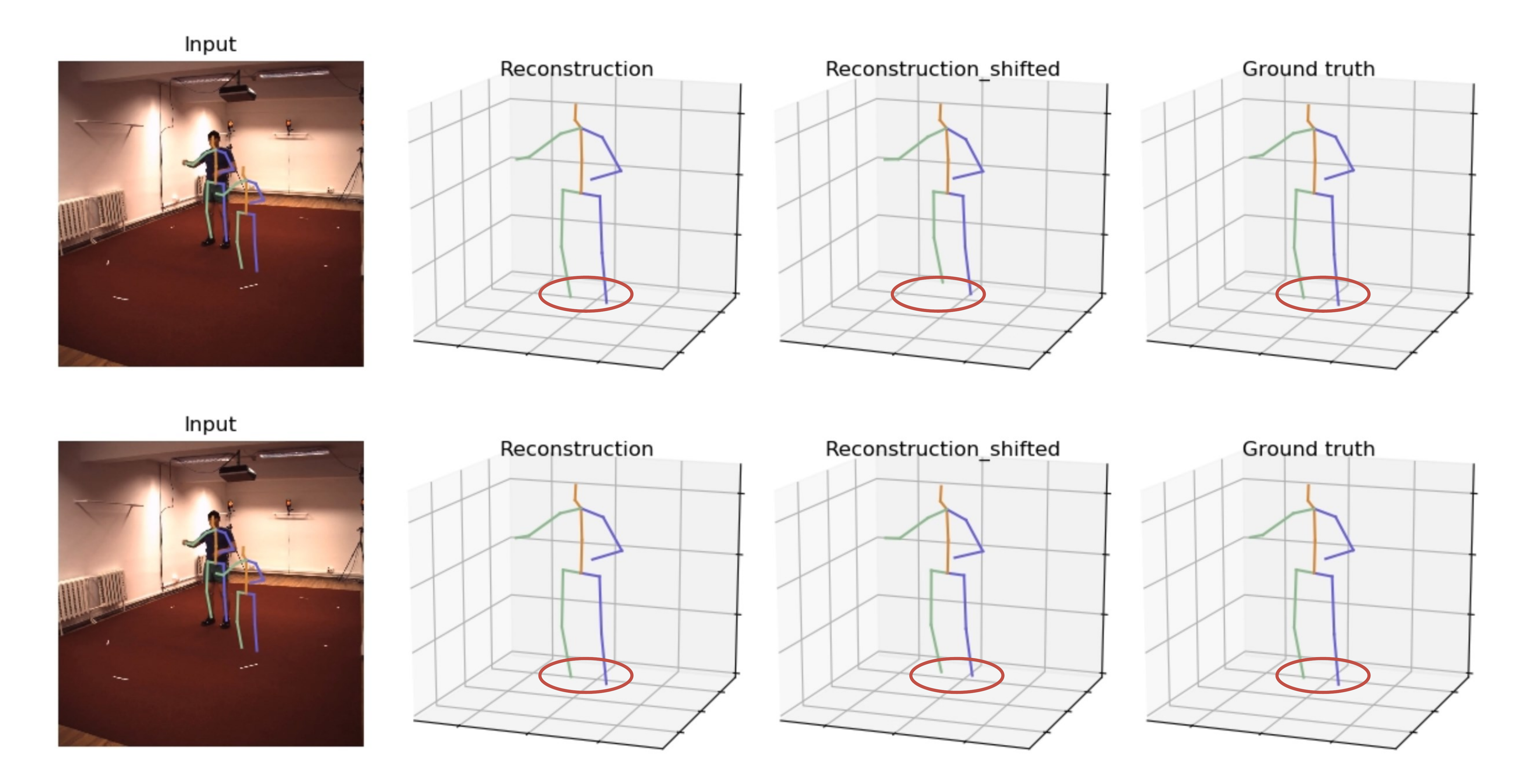}
  \caption{Qualitative results of the baseline and baseline+P-ENH on global motion caused by simulated  camera movement. Both the original and shifted input are drawn in the raw image. Top: baseline. Bottom: baseline+P-ENH.}
  \label{img8}
\end{figure*}

\begin{figure*}[ht]
  \centering
  \includegraphics[width=\textwidth]{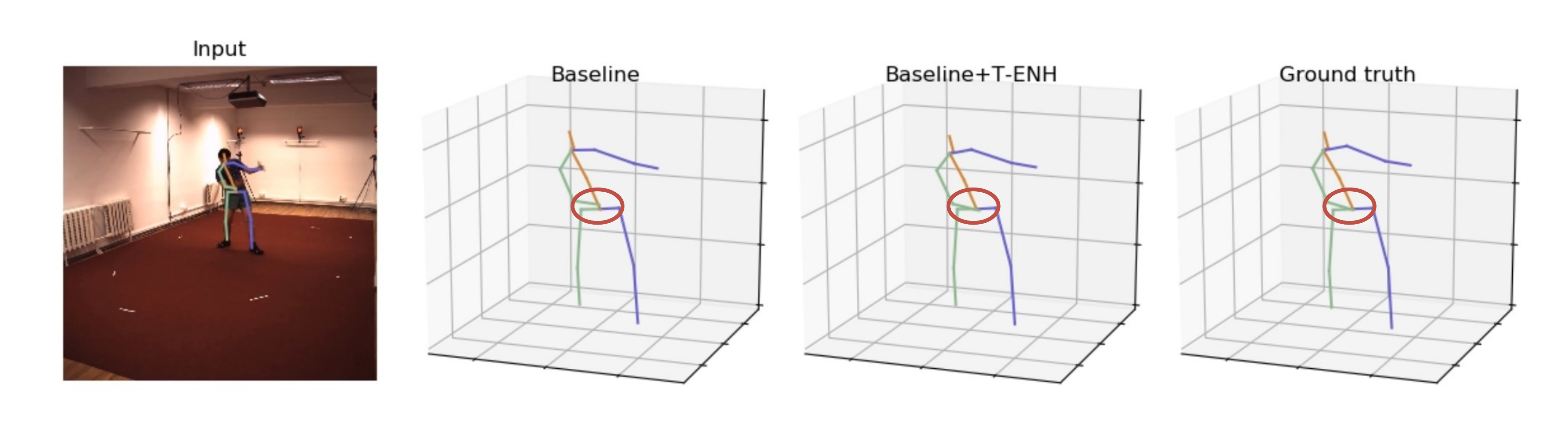}
  \caption{Qualitative results of the baseline and baseline+T-ENH on local motion with a small movement range.}
  \label{img9}
\end{figure*}

\begin{figure*}
  \includegraphics[width=\textwidth]{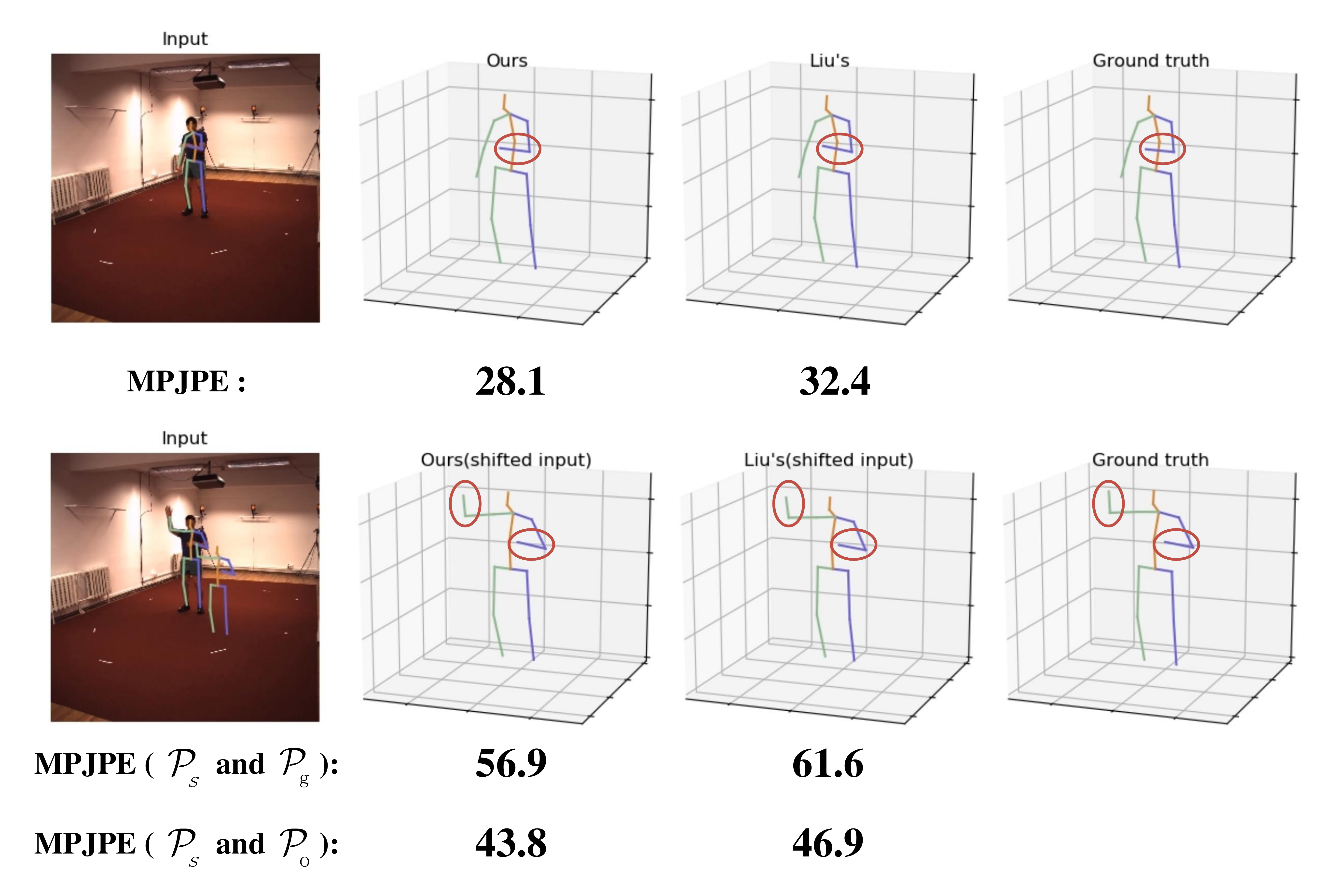}
  \caption{Comparison between our method and Liu's approach~\protect\citesec{liu2020attention_sec}. Top: the case of local motion with a small movement range. Bottom: the case of global motion caused by simulated camera movement. The original and shifted input are drawn in the raw image. }
  \label{img1}
\end{figure*}

Figure~\ref{img8} shows the results of the baseline and baseline+P-ENH on global motion caused by simulated camera movement. When the input is shifted, the baseline+P-ENH produces a result closer to the 3D pose estimated from the original input than the baseline in the foot region. Figure~\ref{img9} shows the results of the baseline and baseline+T-ENH on local motion with a small movement range. It can be seen that the baseline+T-ENH produces a more accurate result than the baseline in the hand region. Figure~\ref{img1} shows the results of our method and the approach carried out by Liu \textit{et al.} \citesec{liu2020attention_sec}. The magnitude of the offset vector is 0.282. Compared to Liu's approach, our method obtains better consistency between $\mathcal{P}_{s}$ and $\mathcal{P}_{o}$ by means of positional information encoding. Besides, our method achieves a higher prediction accuracy on small movements using temporal information encoding.

\section{Implementation Details}\label{implement}
The proposed method is implemented using PyTorch \citesec{paszke2017automatic}. For training and testing, we set the length of the pose sequence $T=243$, which is the same as \citesec{jllo20193d_sec}. We use the proposed multi-stage optimization method to separate the optimization process of the encoders and FFM to ensure that the feature extraction is not affected. In the first stage, Feature Fusion Network without the FFM is trained for 80 epochs with the initial learning rate set to $1e^{-3}$. In the second stage, the parameters of the encoders are loaded from the first stage and fixed. Only the FFM and the decoders are trained for another 80 epochs with the initial learning rate set to $1e^{-3}$. In the third stage, the whole framework is finetuned for 20 epochs with the initial learning rate set to $5e^{-4}$. The learning rate decreases by 5\% after each epoch for all stages. Leaky ReLU \citesec{xu2015empirical} is adopted as the activation function. We adopt 1024 as the batch size. The channel dimension and dropout rate of the TCN are set to 512 and 0.2, and those of the global feature encoder, decoders, and fusion block in the FFM are set to 1024 and 0.25 respectively. AdamW \citesec{loshchilov2018fixing} is used as the optimizer. The dimension of the local, global, and fused features is set to 512. All experiments were conducted on a single Nvidia GeForce GTX 3080 GPU.

\section{Analysis on absolute 2D positions}
We conduct experiments to analyze the effect of absolute 2D positions on the input side. The results are shown in Table~\ref{tab6}. It is found that the absolute 2D positions of joints are necessary to the inference of 3D poses. In the case of positional information encoding, discarding absolute 2D positions will result in the loss of some visual priors, for example, the lower part of the image usually corresponds to the area closer to the camera, and the upper part corresponds to the area farther away. The lack of these priors will affect the depth prediction of each joint, and lead to a sharp performance drop. In the case of temporal information encoding, only retaining the temporal enhanced representation will cause the input of the current pose to be 0. Therefore, we utilize the absolute 2D position in the current frame as well as the temporal enhanced

\begin{center}
    \vspace{0.3cm}
      \captionof{table}{The effect of absolute 2D positions on the prediction results. w/o ABS: without absolute 2D positions}
    \begin{tabular}{ccc}

    \toprule
    Method & MPJPE(mm)  & $\Delta$ \\
    \midrule
    Baseline & 34.1&-  \\
    +P-ENH & 33.4&0.7  \\
    +P-ENH(w/o ABS) & 36.2&-2.1  \\
    +T-ENH & 33.1&1.0  \\
    +T-ENH(w/o ABS) & 36.9&-2.8  \\

    \bottomrule
    \end{tabular}
    \vspace{0.5cm}
  \label{tab6}
\end{center}

\begin{center}
\centering
  \captionof{table}{Comparison between different temporal information encoding operators.}
    \begin{tabular}{ccc}
    \toprule
    Method & MPJPE(mm)  & $\Delta$ \\
    \midrule
    Baseline & 34.1&-  \\
    +CP & 33.9&0.2\\
    +IP & 33.6&0.5\\
    +CS & 33.5&0.6\\
    +SUB & 33.1&1.0  \\
    +SUB+SUB\_S & 34.2&-0.1  \\
    +SUB(81f)& 33.7&0.4 \\
    \bottomrule
    \end{tabular}
     \vspace{0.5cm}
  \label{tab5}
\end{center}

\begin{center}
\centering
  \captionof{table}{The influence of the M-S OPT method on P/T-ENH.}
    \begin{tabular}{ccc}
    \toprule
    Method & MPJPE(mm)  & $\Delta$ \\
    \midrule
    Baseline & 34.1&-  \\
    Baseline(M-S OPT) & 33.3&0.8  \\
    \midrule
    +P-ENH & 33.4&-  \\
    +P-ENH(M-S OPT) & 32.5&0.9  \\
    \midrule
    +T-ENH & 33.1&-  \\
    +T-ENH(M-S OPT) & 31.8&1.3  \\
    \midrule
    +P-ENH+T-ENH &32.3&- \\
    +P-ENH+T-ENH(M-S OPT)&30.1 & 2.2\\
    \bottomrule
    \end{tabular}
    \vspace{0.5cm}
  \label{tab7}
\end{center}

\noindent representation in other frames as inputs. However, this will lead to another problem that the object of the convolution operation includes both the absolute 2D position and temporal enhanced representation. The scale of these two is different, which brings a burden to the learning process of the network and deteriorates the prediction performance. In summary, the absolute 2D position is indispensable, so we use the original input together with the positional and temporal enhanced representations as the inputs to the Feature Fusion Network.

\section{Choices for Temporal Information Encoding Operator} We evaluate the performance of different temporal information encoding operators in Table ~\ref{tab5}. The notations are defined as:
\begin{itemize}
 \item \textbf{CP}: cross-product.
 \item \textbf{IP}: inner-product.
 \item \textbf{CS}: cosine similarity.
  \item \textbf{SUB}: subtraction.
  \item \textbf{SUB+SUB\_S}: the geometric descriptor used in \citesec{yang20183d}. It consists of the differences between the current pose and the others as well as the quadratic deformation of the differences.
  \item \textbf{SUB(81f)}: only use the differences in 81 frames around the current pose (243 frames in total).
  
\end{itemize}
It is shown that subtraction is a more appropriate operator than cross-product, inner-product, and cosine similarity. The long-term dependencies will not have a negative impact on the prediction of the current pose as the performance of SUB is better than SUB(81f). Additional information, like the quadratic deformation of the difference, is not needed. Therefore, we choose subtraction as the temporal information encoding operator in our paper.

The intuition why subtraction is a relatively good choice is as follows. Since 3D pose estimation from 2D poses is an ill-posed problem due to the inherent ambiguity in depth, perceptual evidence needs to be added to assist the estimation of depth. Previous work \citesec{rogers1979motion} shows that motion parallax provides a reliable cue for monocular depth perception. The differences between the current pose and the others in the time domain play a similar role as the motion parallax. They are beneficial to estimate the depth of each joint under monocular conditions, and help the networks converge to a better result.

\section{Discussion on the effect of different components on the performance}
As analyzed in the ablation studies, P-ENH, T-ENH, and M-S OPT reduce the error by 0.7mm, 1.0mm, and 2.2mm respectively. The reason why the M-S OPT method has the strongest effect on the performance is as follows. M-S OPT is designed for the overall framework, so it brings a relatively large performance improvement. P-ENH and T-ENH are proposed to alleviate two different problems in specific scenarios. The performance improvement brought by P-ENH is secondary because it mainly focuses on improving the robustness to global motion. T-ENH focuses on improving prediction accuracy on small movements but has a limited effect on large movements. The test set includes both large and small movements, so the reduction in the mean error is not as significant as that of the experiments conducted on small movements.

Besides, we analyze the effect of the M-S OPT method on P/T-ENH. As shown in Table~\ref{tab7}, we apply this method to the baseline, baseline+P-ENH, baseline+T-ENH, and baseline+P-ENH+P-ENH. The M-S OPT method improves upon the baseline by 0.8mm, while improves upon the baseline+P-ENH and baseline+T-ENH by 0.9mm and 1.3mm. This shows the effectiveness of M-S OPT in exploiting the positional and temporal information.


\bibliographystylesec{ACM-Reference-Format}
\balance

\end{multicols}

\end{document}